# MedImgAnnotator: batch wise medical image annotation and preprocessing tool

Ewunate Assaye Kassaw[1,2], Amit Mehndiratta[1,3,4,5], Peush Sahni[6], Anup Singh[1,3,4]

[1]Centre for Biomedical Engineering, Indian Institute of Technology Delhi, New Delhi, India

[2]Department of Biomedical Engineering, University of Gondar, Gondar, Ethiopia,

[3]Yardi School of Artificial Intelligence, Indian Institute of Technology Delhi, New Delhi, India

[4]Department of Biomedical Engineering, All India Institute of Medical Science, New Delhi, India

[5]Faculty of Medicine and Health, University of New South Wales (UNSW), Sydney, Australia

[6]Max Centre for Liver and Biliary Sciences, Max Super-specialty Hospital, Saket, New Delhi, India.

anupsm@iitd.ac.in

**Abstract:** The curation of large cohort datasets for developing deep learning models demands annotating or editing over the AI-predicted labels. However, openly available annotation tools are time-consuming, as they involve complex navigational steps and do not support batch processing. This study aimed to develop a simple, fast, semiautomated batch-annotation tool named MedImgAnnotator. The tool was created by integrating editing tools, image mask pair matching logic, preprocessing functions, object detection tool, and a plugin management system. Its performance was evaluated in terms of active annotation time and ease of use by the developer and 12 users using open-source MR and CT scans. Additionally, the tool had been used by the MICCAI 2026 CARE Challenge liver segmentation best performance winner to annotate liver masks from 6412 MRI volumes. The evaluation results showed that MedImgAnnotator reduced active annotation time from 206.5±32.87 (average active annotation time of openly available tools) seconds to 59 seconds. The Wilcoxon signed-rank test of 12 users' evaluation results recorded P values of 0.032 and 0.008 for the active annotation time and higher ease of use rating metrics, respectively, indicating that the current tool significantly reduces annotation time and enhances ease of use. Moreover, the tool's utilization for annotating different regions of interest from different modality scans and body regions shows its usability across modalities and areas of interest. MedImgAnnotator, which significantly reduces active annotation time for large imaging cohorts, is available as an executable file and as a Python script. access link

## 1. Introduction

Segmentation of medical images is important in diagnosis and treatment planning/monitoring, facilitating objective, and quantitative evaluations for clinical decision-making[1, 2]. Recent advancements in deep learning-based image segmentation algorithms have significantly improved the speed and accuracy of automated medical image analysis[2, 3], with foundation models, such as the Segment Anything Model (SAM) and its medical adaptations (MedSAM) [4–7], have demonstrated promptable zero-shot segmentation capabilities across diverse imaging modalities. However, despite this generalizability, these models remain fundamentally limited for practical clinical ground-truth curation to unstable and error-prone nature of the zero-shot segmentation for domain specific and ambiguous tasks. Furthermore, these models demand high performance computational resource and complex environmental setups. As a result, fulfilling the massive, annotated data demands of the supervised and domain specific accurate models remains highly challenging as their performance is intrinsically dependent on data quality and volume. [2, 8]. Usually, clinical experts carefully annotate these ground-truth segmentations to delineate Region of Interest (ROI) boundaries. Currently there are only few open access noncommercial image annotation and processing tools for medical imaging data; such as 3D Slicer [9], ITK-SNAP [10, 11], ImageJ/Fiji [12], MITK [13], and MRIcron [14].

Although these existing tools provide robust image processing engines, their user interaction paradigms were primarily designed for single-volume exploratory visualization rather than high-throughput batch dataset annotation. 3D Slicer software is widely used for medical image filtering, segmentation, registration, and 3D volume rendering. It is very adaptable, accommodating several plugins for expert researchers. However, its broad reach makes it very resource-intensive and computationally demanding, sometimes leading to a steep learning curve and a complex interface for those that need merely rapid annotation capabilities[9]. ImageJ/Fiji is a multifaceted software designed for performing several image processing operations, including stitching, distortion correction, and elastic alignment, etc. However, image annotation on multiple slices and multiple volumes is tedious and time-consuming task. While deep learning plugins may be included into Fiji, it is mostly optimized for 2D microscopy and biological imaging, and it lacks effective native volume management for rapid clinical batches in 3D[12]. ITK-SNAP is known for its user-friendly interface, facilitating the viewing, alteration, and annotation of images using dynamic contours and three-dimensional region expansion techniques. Although ITK-SNAP offers sophisticated semi-automated tools, it necessitates users to manually upload images and masks individually, rendering mass processing quite tedious[10, 11, 15]. MRIcron (and its successor MRIcroGL) are lightweight imaging tools that handle different data formats and allow basic markups using pen and filling options; however, they lack advanced shape-based interpolation and auto-snapping features[14]. Finally, MITK provides a robust framework for integrating image processing algorithms, but like 3D Slicer, it suffers from software heaviness and complex multi-step user interfaces[13].

These widely used tools remain fundamentally architected for single-volume exploration visualization rather than high-throughput batch data annotation. Moreover, the tools don't support object detection annotation. Thus, the availability of an annotation and preprocessing tool that performs automated case queuing, slice interpolation, and lightweight offline deployment remained the major gap. This work was motivated by a clinical research project that required ground-truth mask annotations for over 14,270 multi-parametric and multi-phase abdominal MR and CT volumes. In a pilot study, annotation of 50 liver CT volumes using 3D Slicer took around 24 hours due to individual volume file loading, manual volume-mask alignment, and multi-tier menu navigation. This study aims to develop and evaluate MedImgAnnotator, a semi-automated batch annotation and preprocessing tool that reduces active annotation time and physical user engagement while preserving the demanded annotation region of interest.

## 2. Methods

In this study a software named MedImgAnnotator was developed for simplifying and facilitating bulk image annotation and preprocessing of medical images. Initially the functional requirements, such as the brush, eraser, image mask matching logic, 3D interpolation, and other important functions for performing medical image annotation, were identified. Image display, scrolling, zooming, etc. capabilities were also identified. Then the identified functions were implemented using the Python programming language and their corresponding algorithms. Consecutively, all the implemented functions were integrated into a unified graphical user interface (GUI) and compiled into a standalone application using the PyInstaller Python library and the Inno Setup compiler. The plugin manager is incorporated to allow the users to integrate any image processing modules and AI models. Performance evaluation and comparison against the commonly used openly available annotation tools were performed, followed by an iterative optimization and refinement. Finally, usability tests and comparative evaluation were performed by 12 users and made the free trial version of the compiled tool accessible. Figure 1 illustrates the overall development pipeline used in this study. Further details are provided in the following sections.

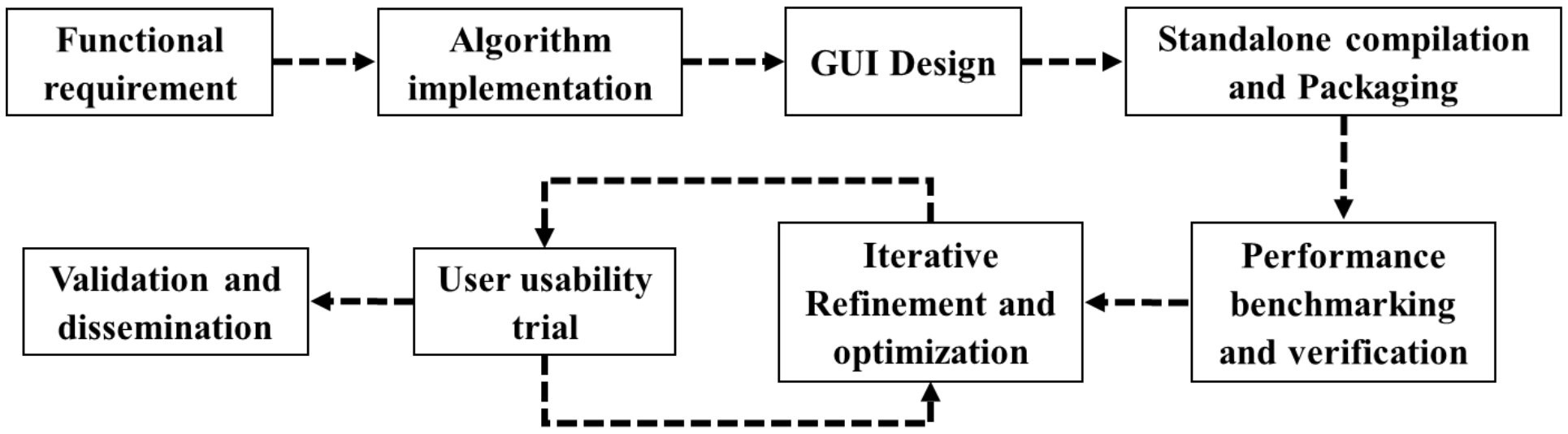

Figure 1: MedImgAnnotation tool development procedure

### 2.1. Data Information

This study utilized publicly available and anonymized chest CT images from the chronic heart diseases dataset of MICCAI 2019 [16, 17], abdominal MR and CT images from Abdominal Multi-Organ Segmentation (AMOS 2022) [18], brain MR images from the 3D slicer tool brain tumor sample data [9], and abdominal MR images from the Pancreas MR dataset [19] for testing the tool's annotation performance and usability. Moreover, this tool was utilized for preparing and annotating the dataset utilized in the iterative human-in-the-loop liver segmentation study [20].

### 2.2. Software distribution and environment

The MedImgAnnotator software was developed in Python (version 3.12.7, 64-bit) with CustomTkinter (v5.2.2) for the graphical user interface, NumPy (v1.26.4) and SciPy (v1.16.0) for array calculations, and NiBabel (v5.3.2) and SimpleITK (v2.5.2) for medical volumetric image processing and affine registration. The image processing and morphological refinement were performed using scikit-image (v0.25.2) and OpenCV (v4.12.0). All computational tasks and evaluations were performed on a host workstation with a 12$^{th}$ Generation Intel® Core™ i7-12650H (10 cores, 16 threads, base frequency 2.30 GHz), 16 GB DDR4 RAM (3200 MHz), and an NVIDIA® GeForce RTX™ 2050 GPU running Windows 11 64-bit. The tool was evaluated by different users on different desktop computers and laptops with varied configurations.

### 2.3. Functional requirement

These days, as deep learning models assist medical image processing researchers and require a large cohort of data, editing the labels based on the initial predictions simplifies the overall annotation process. To do so, bulk image and initial prediction matching, as well as color overlaying the masks on their corresponding base images, are vital tasks. Overlaying the images with their corresponding mask is a vital task for simplifying the annotation process. Both editing the overlaid mask and annotating from scratch using brush, bucket, freehand poly, wand, and eraser functions play a significant role in performing fundamental annotation activities. Interpolation plays a significant role in minimizing the required annotation time by interpolating the structures on intermediate slices from the annotated slices. The undo and save functions are important for moving back and saving the final annotation results, respectively. The image preparation module facilitates the preparation and processing of the data prior to performing annotation. The plugin manager helps the user integrate necessary external functions into this tool.

### 2.4. Algorithmic implementation and GUI design

#### 2.4.1. Data management, matching, and overlaying

The initial components of the tool select data path, maps the volume's X, Y, and Z axes to the axial, sagittal, and coronal display planes, and overlays the available masks on the base images. The proposed tool takes image data input in NIFITI format. Data from multiple patients or volumes to be annotated need to be placed into same directory. The directory traversal with sorting functions were used to select all the NIFTI files in the given path and sort them alphabetically and give reproducible list every time the user selects the same path. Then the setting class was developed to save the checkpoint (the last opened image key) for resuming the operation later from the point the user left off. The function designed for matching the image-mask pairs utilizes exact file name matching, numeric token matching, and approximate string-matching approaches to match the masks with their corresponding image if the mask path is given. Initially it starts with the exact file name matching, if it doesn't work it will look for the numeric token matching and if again the numeric token is not working the approximate string matching will be used. The approximate string-matching technique utilizes Ratcliff–Obershelp (Gestalt) pattern-matching algorithm [21] with the formula given in the equation 1.

$$ratio = \frac{2M}{|a| + |b|} \dots\dots\dots\dots\dots\dots\dots\dots\dots\dots\dots\dots\dots\dots\dots 1$$

Where $M$ is the total number of matching characters in the longest contiguous match found between the strings 'a' and 'b' iteratively matching files. The matching will be done by setting the threshold value of 0.6, the value below 0.6 is considered as not matching.

#### 2.4.2. Volume loading, Intensity adjustment, and display

Functions from 'Nibable' library in Python were used to load NIfTI image volumes from the prepared batch queue while preserving spatial affine transformations and header metadata. To support both high-speed display rendering and quantitative precision, a 32-bit floating-point copy (float32) of the raw physical voxel intensities is retained in background memory for intensity-guided algorithms, while linear min–max normalization [22] given in equation 2 was used to convert the display image volume directly into an 8-bit unsigned integer representation.

$$I = \left[\frac{I - I_{min}}{I_{max} - I_{min}}\right] \times 255 \dots\dots\dots\dots\dots\dots\dots\dots\dots\dots\dots\dots\dots\dots 2$$

Where I, $I_{min}$, and $I_{max}$, represents the voxel intensity, the minimum intensity of the image, and the maximum intensity of the image, respectively.

Loading an initial or predicted segmentation mask is completely optional. While annotating cohorts from scratch without prior deep-learning predictions, an empty 8-bit unsigned integer (uint8) mask array matching the base image dimensions is automatically initialized in memory without raising load errors. When pre-existing or predicted masks are provided for refinement, the matched mask volume is loaded, rounded to the nearest integer, cast to an 8-bit unsigned integer (uint8), aligned with the base image voxel grid, and overlaid on the base images. NumPy slicing is used to extract the active 2D orthogonal slice from the 3D volume, followed by coordinate reorientation for screen display. Affine intensity transformation [23–25] is used to apply interactive contrast and brightness windowing via a precomputed 8-bit Look-Up Table (LUT), where the transformed display intensity I' is scaled in the range of [0, 255] using the formula given in Equation 3. .

$$I' = clip(c.I + b, 0, 255) \dots\dots\dots\dots\dots\dots\dots\dots\dots\dots\dots\dots\dots\dots 3$$

Where c and b denote the contrast and brightness parameters, respectively, which the user adjusts with scroll bars. The 8-bit gray scale image slice is composited with the label color using an over operator of alpha compositing [26], resulting in the blended output value given by Equation 4. The compositing is applied through vectorized Boolean masking, so only pixels belonging to the labeled structure are blended with the label color at a user-selectable opacity a ∈ [0, 255], while retaining unlabeled pixels to their original image intensity.

$$C_{out} = \frac{I(255 - a) + C.a}{255} \dots\dots\dots\dots\dots\dots\dots\dots\dots\dots\dots\dots\dots\dots 4$$

The largest scale factor is used to keep the selected 2D slice of the image within the panel bounds and fit the image into the viewport while keeping the aspect ratio [27]. The display size will be recomputed every time the user resizes the application window. Additionally, the vertical scroll bar allows the user to zoom in and out on the displayed 2D slice. Bilinear interpolation is used to recompute the display during up sampling (zooming in) [28]. Anti-aliased area filtering is used for down sampling (zooming out) [29]. Nearest-neighbor interpolation [30] is used for up sampling and down sampling the labels overlaid on the 2D slice image while preserving the discrete label values and boundaries. 2D image translation [28] by a displacement vector v is applied through the mouse drag action, which moves the image within the viewport.

### 2.4.3. Annotation functions

The 2D circular or the 3D spherical brush, the 2D circular or the 3D spherical eraser, the flood-fill bucket, the free-hand polygon drawer, the intensity and bounding box guided region growing tool, and the hybrid deformable registration and signed distance transform based slice interpolation were the six core interactive annotation tools of MedImgAnnotator. A core function was implemented to take any mouse event on the 2D screen, which is the user input to the annotation tool, and map it to the pixel grid of the corresponding image. It recovers the selected screen pixel coordinate and then inverts the vertical axis to match the image coordinate convention. This function is the foundation for writing every mouse-driven change on the screen into the mask. The interactive tools built on it are the circular/spherical brush, the circular/spherical eraser, the flood-fill bucket, the freehand polygon drawer, and the intensity- and bounding box-guided region growing tool. The hybrid deformable registration and signed distance transform-based slice interpolation tool then fills the unannotated slices between the manually annotated key slices. An undo function stores a copy of the mask before each edit on a stack, and each undo action pops the most recent saved mask and restores it.

#### 2.4.4. Brush, painting and erasing

The annotation tool includes 2D circular and 3D spherical brush-based painting functions for generating masks for 2D slices and 3D volumes respectively. The brush labels every pixel (in the case of 2D slices) or voxel (in the case of 3D volumes) that falls inside its radius, centered on the cursor position, with the selected label. Pixels outside the brush are left unchanged.

The user can adjust the thickness of the brush by changing the radius. By default, the brush acts as a painting tool. If the eraser is selected the brush will play exactly the opposite role of the painting which is unlabeling all the voxels or pixels located within the radius range. Tkinter [31] function was used to convert the continuous mouse dragging events into sequence of discrete motion events and labeling the pixels or voxels in the distance range to the intended label. Chebyshev distance technique [32] was used to avoid the gaps between the two sampled cursors in the case of fast mouse dragging. This technique helps to interpolate multiple brush centers between the given selected end point brush centers resulting in smooth overlapping brush spheres or circles that gives smooth continuous and uniform thickness labeling while dragging the cursor.

#### 2.4.5. Free hand polygon

A free hand polygon was designed to overcome the time taking and tedious process of painting larger and irregular regions one by one using brush. In this tool a function that samples the points as the user drags the cursor based on the Euclidian arc length [33] was defined. For a sequence of raw cursor positions $P = \{p_1, p_2, \dots, p_M\}$, the total Euclidean arc length $L$ is calculated by equation 5 and was used to recompute uniformly spaced control points throughout the cursor trajectory to balance the dragging speed variations. The new control points or vertices can be calculated by $q_k = p_i + t(p_{i+1} - p_i)$ where $t = \frac{d_{target} - d_{current}}{L_i}$.

$$L = \sum_{i=1}^{M-1} || p_{i+1} - p_i ||_2 \ldots\ldots\ldots\ldots\ldots\ldots\ldots\ldots\ldots\ldots 5$$

Consequently, Chaikin's algorithm [34] was used to get smoothed evenly spaced points in the path of the cursor by redefining the vertexes $(Q, R)$ using the formula given in equation 6.

$$(Q, R) = (\frac{3}{4}P_i + \frac{1}{4}P_{i+1}, \quad \frac{1}{4}P_i + \frac{3}{4}P_{i+1}) \ldots\ldots\ldots\ldots\ldots\ldots\ldots\ldots\ldots\ldots 6$$

Upon the formation of the closed polygon ray casting algorithm [35] was used to label every pixel grid in the drawn polygon as a desired label and exclude the pixels outside the polygon.

### 2.4.6. Region growing and automated edge snapping functions

The bucket tool uses connected component labeling technique [36] to label the pixels (fill areas) enclosed by existing labels. On the contrary, the wand tool uses a bounded seeded region-growing algorithm that directly applies to the image. Upon clicking on the image, the algorithm samples the local intensity and uses it to label the pixels or voxels that have a similar intensity profile to the seed, based on user-adjustable radius limit (Lim) and threshold tolerance (Tol) values. The automatic edge snapping function uses an image foresting watershed [37, 38] algorithm to refine the contour via morphological erosion and dilation across the local intensity gradient magnitudes.

### 2.4.7. Slice Interpolation

Multi-slice interpolation is implemented using a hybrid framework that combines Fast Symmetric Forces Demons deformable registration with Signed Distance Transform (SDT) shape interpolation [39–44] to accurately track translating, curving, and deforming 3D anatomical structures across slices while preserving smooth boundaries and preventing ghosting artifacts.. It converts the reference masks into SDT given by, $SDT(x) = DT_{int}(x) - DT_{ext}(x)$. For a starting slice $z_1$ and ending slice $z_2$, the intermediate distance field at a target slice $z$ is calculated using a spatial relative weight $w$ given by equation 7 if the keyframe masks are nearly stationary (Dice >= 0.98).

$$\Phi_{interp}(x) = (1 - w)SDT_1(x) + wSDT_2(x) \ldots\ldots\ldots\ldots\ldots\ldots\ldots\ldots\ldots\ldots 7$$

If the keyframe masks are not stationary that happens when anatomical structures shift laterally or deform across slices (Dice < 0.98), adjacent grayscale slices within a padded region of interest are pre-aligned via 2D Normalized Cross-Correlation (NCC) and registered using multi-scale coarse-to-fine Fast Symmetric Forces Demons optical flow. The reference distance fields are warped bi-directionally along the resulting inter-slice displacement fields, corrected via co-moving centroid residual transport, and fused using narrow-band NCC confidence weights to yield $\Phi_{interp}$. A Gaussian filter [24, 45] ( $\sigma = 0.8$ ) is applied to $\Phi_{interp}$ to ensure biologically organic curves, and the final intermediate mask is extracted by thresholding the field where $\Phi > 0$. If auto-snapping is enabled, morphological operations and IFT are applied to align the mathematically guessed shape to the nearest local image edges.

### 2.4.8. Object detection

A function for drawing bounding boxes with class labels is included for annotating 2D and 3D datasets for object detection projects. Selecting the object detection (ObDet) button shifts the tool from the segmentation annotation tool to the object detection bounding box drawing tool and automatically creates a bounding box of 64 × 64 pixels at the center of the opened image. Zooming will be fixed to 1.0, which allows the direct pixel mapping while changing the mouse cursor position. Clicking near the corner of the existing box allows the user to resize it, while clicking inside the box allows the user to drag it to the required position. Pressing delete or backspace keys allows the user to remove the selected box. Scrolling the mouse wheel allows you to zoom the bounding box. Finally, the prepared bounding boxes will be saved either by pressing the S key or clicking on the save button.

### 2.4.9. Image preparation and processing package

To enable an end-to-end standardized pipeline for medical imaging curation, annotation, and deep learning model training, a comprehensive image preparation, preprocessing and postprocessing toolbox (ImPrep) was integrated into the primary graphical user interface of the MedImgAnnotator application immediately upstream of the data input pipeline. The integrated suite serves as a unified environment for preliminary data cleaning, parameter optimization with real-time visual inspection, and batch cohort processing. The toolbox incorporates comprehensive modules for data standardization and conversion (DICOM-to-NIfTI with series and echo-time sorting, 2D slice stacking, lossless compression, and bit-depth conversion), spatial harmonization (canonical RAS reorientation, affine header correction, 4D channel decomposition, and multi-dimensional voxel resampling), and intensity processing (CT Hounsfield Unit windowing, contrast enhancement, and multi-scheme statistical normalization). In addition, segmentation thresholding, 3D connected-component labeling, and dataset curation utilities (modality sorting, name similarity-based filename matching, and reproducible train–validation–test cohort splitting) are incorporated. By embedding these algorithms directly within MedImgAnnotator, the platform provides a cohesive workflow that bridges raw multi-modal imaging data, manual and semi-automated ground-truth annotation, and input preparation for deep learning model development.

### 2.4.10. Extension Plugins

A self-contained method was defined to handle the plugin independently without incorporating it in the main tool. This plugin manager method will facilitate the installation of plugins that are prepared in form of python script, folder containing multiple scripts, or zipped folder/file, displaying and running the available installed plugins, and accessing the plugin source scripts from the plugin installation folder. This method places the plugin script/s into the designated folder for the plugin, and the required dependencies and libraries will be installed while installing the plugin itself. For this reason, every plugin must be prepared in such a way that it must install all the required dependencies upon its first ever running. Upon the first ever run of the plugin it identifies the required packages, attempt to import, and records the failing import as the missing list. The subprocess method of the plugin is used to install the missing libraries from the missing list obtained during the first run of the plugin.

#### 2.4.10.1. Automatic deep learning-based segmentation plugin

A plugin was prepared for running the developed nnunet-v2 [46] liver segmentation model cascaded with the 3D connected component filtering method. Mini-window user interface was designed that will be used for the user to insert the input directories and visualize the segmentation progresses. Upon the first ever run of the plugin it identifies the PyTorch and nnU-Net v2 packages which are the backbone libraries for running the developed model were missing. Consequently, the subprocess method installed the PyTorch and nnU-Net v2 packages. After installing the dependencies, the plugin will be able to segment the liver from abdominal MR scans found in the given input folder and saves the output into the user selected output folder.

### 2.5. Software Packaging, Distribution, and System Requirements

MedImgAnnotator was compiled and packaged into a self-contained, standalone Windows desktop application using PyInstaller. We used directory-based bundling to avoid runtime extraction latency. The application was packaged into a turnkey installer (.exe) using Inno Setup which sets up file associations and desktop shortcuts. MedImgAnnotator requires a 64-bit Windows operating system with a minimum of 8 GB RAM 2.0 GHz multi-core CPU and 5 GB free disk space for optimal performance\. The core application does not require any external python runtime, cuda installation or compiler toolchain. Optionally, an NVIDIA GPU can be used for hardware-accelerated deep learning inference via the nnU-Net plugin. The Supplementary Information contains the standalone installer executable and a detailed written User and Installation Manual (PDF) that is permanently archived for the reader to download.

### 2.6. Performance Evaluation and refinement techniques

The main comparative performance metric in this study was the active annotation time. Active annotation time was measured as the total active time to open a case, verify/align initial masks, edit erroneous boundary labels, and save the finalized segmentation mask. The active annotation time (the elapsed time while performing) was measured in seconds. The user-perceived simplicity and workflow satisfaction rating was used as a secondary evaluation metric. The user rating was measured on a validated 5-point Likert scale from 1=highly cumbersome to 5=exceptionally efficient.

The tool's performance evaluation was conducted in three phases. In the first phase, the liver parenchyma was annotated using MedImgAnnotator, 3D Slicer, ImageJ, ITK-SNAP, and MRIcron from the same volume. The number of seconds that the annotation took were recorded for each tool. After the first phase of testing, the tool was extensively used to prepare and annotate 7093 multi-parametric, multi-phase, and multi-centric abdominal MR scans (pseudo labels were annotated and refined using this tool for the unlabeled 6412 volumes) to develop a generalizable liver segmentation model as part of the MICCAI CARE Liver 2026 challenge, which secured the first rank on the leaderboard of the liver segmentation challenge[20]. Multiple refinements were made based on the comments received from colleagues using the tool and observations from developer.

In the second phase, a comparative usability test was performed by 12 researchers with 1-5 years of experience in image processing who annotated their region of interest using MedImgAnnotator and one of the four image annotation tools from 3D Slicer, ImageJ, ITK-SNAP, and MRIcron from the same volume. For uniformity and avoiding any bias, the user training was provided prior to the annotation for MedImgAnnotator, on 5 datasets. The users rated the usability and simplicity of the tool out of five in addition to recording the number of seconds that the annotation took for them. In the third phase the liver, brain tumor, lung, pancreas, spleen, and kidneys were annotated from 20 abdominal CT and MR volumes, 5 brain MR scans, 5 chest MR scans, and 5 abdominal MR scans, respectively. The time required for annotating the region of interest from each volume was recorded along with feedback on usability, ease of the tool usage for performing annotation.

The Friedman statistical test was performed to assess the performance of MedImgAnnotator in terms of the time it takes for performing a certain annotation task against four open access commonly used annotation tools (ImageJ, 3D-slicer, ITK-snap, and MRIcron) from single-user data. Additionally, the Wilcoxon signed-rank test was performed to evaluate the performance of the developed tool against the above four tools using the annotation time record obtained from 12 independent users; each of them used MedImgAnnotator and one of their favorite tools among the four tools. The Mann-Whitney U test was applied on the annotation time recorded while annotating the liver from 20 MR volumes and 20 CT volumes by a single user.  Kruskal-Wallis tests were performed for analyzing the annotation time difference between the annotation of brain tumors from brain MR images and the annotation of lungs from chest MR scans. Finally, the Friedman test was performed to evaluate the significant time difference between annotating the spleen, pancreas, and kidney from the same MR volumes. All analysis was conducted at a statistical significance level of $\alpha=0.05$.

## 3. Results

The total number of volumes used in this study to test MedImgAnnotation was 58. The datasets were 5 brain MR volumes obtained from 3D slicer sample data; 5 chest MR scans were obtained from the MICCAI 2019 chronic heart diseases dataset, 20 MR and 20 CT abdominal scans were obtained from AMOS dataset, 5 abdominal MR scans were obtained from pancreas MR dataset, and three chest CT scan volumes from MICCAI 2019 chronic heart diseases dataset. Additionally, this software was used to edit the deep learning model prediction masks from 6412 MR volumes utilized in the MICCAI 2026 CARE challenge liver segmentation task, which helped assess the practical applicability of the tool and update it accordingly.

### 3.1. User Interface and Functional Workflow

MedImgAnnotator consolidates comprehensive dataset management (directory scanning, case searching, multi-planar orthogonal slicing, window/level contrast adjustment, canvas zooming, pan translation, and one-touch saving) alongside precision editing tools (raster brush, eraser, flood bucket, intensity wand, shape interpolation, and undo/redo stacks) into a unified graphical user interface. Moreover, the Imprep (image preprocessing) module that contains image preparation and preprocessing utilities to prepare and preprocess heterogeneous data to curate a homogenous dataset is incorporated. Figure 2 shows the visual layout, controls, dropdown selectors, and display panels of the tool.

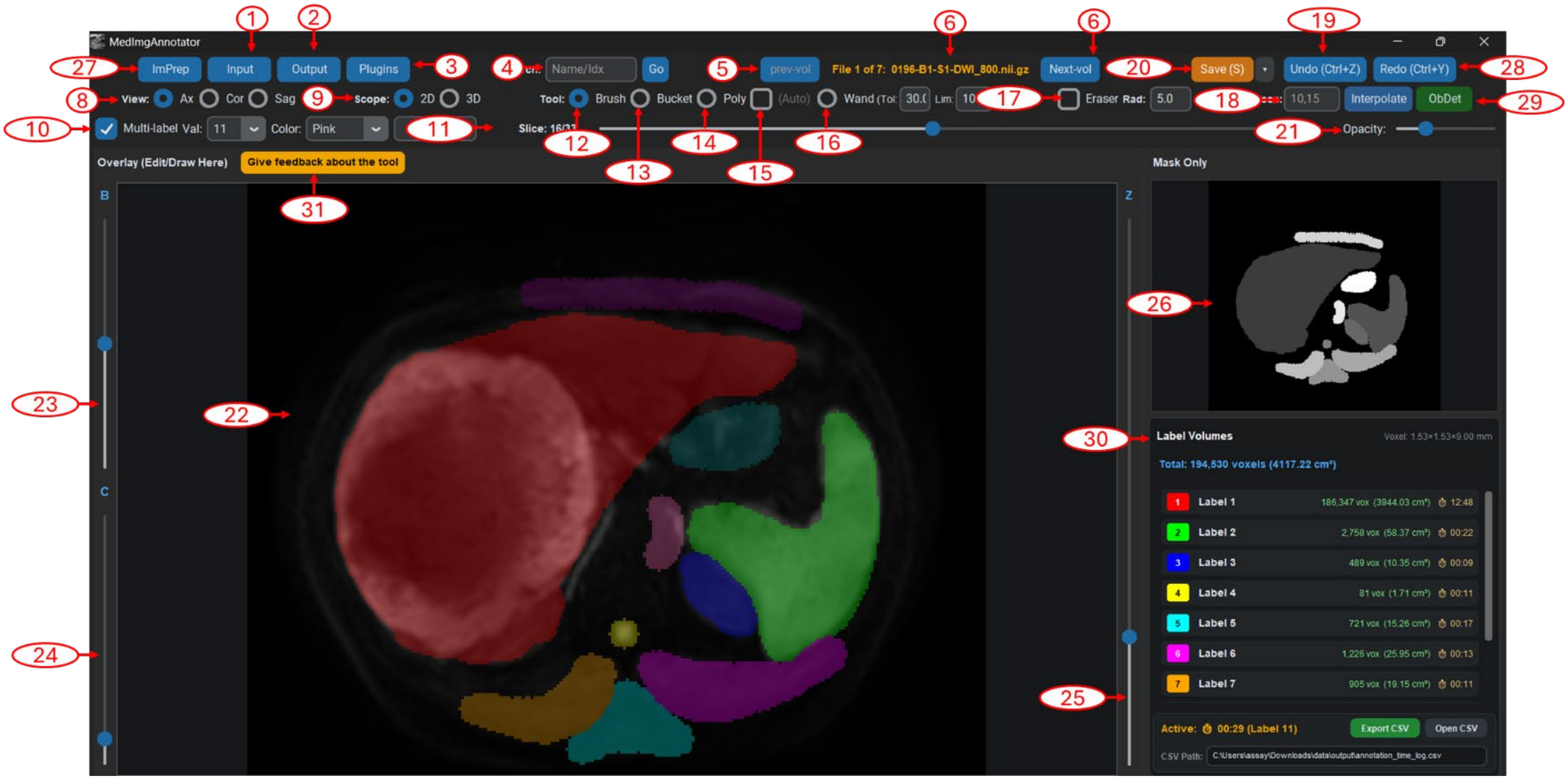

Figure 2: The overview of the developed tool (1:input directory selector, 2:output directory selector, 3:plugin manager, 4:searching bar, 5&7 file Queue navigation buttons, 6:name of the file under process, 8:view selector radio buttons, 9: annotation dimension, 10:multilabel and single label selection option, 11:slice changing scroll bar, 12:brush tool, 13:bucket tool, 14:polygon drawing tool, 15:automatic polygon edge snapping tool, 16:automatic area snapping tool, 17:erasor selection radio button, 18:slice range selection and execution button, 19:undo, 20:single click save button, 21:overlaid mask opacity scroll bar, 22:overlaid image mask display panel, 23:image brightness adjustment scroll bar, 24:image contrast adjustment scroll bar, 25:image zooming scroll bar, 26:mask display panel, 27:image preparation and preprocessing toolbox, 28:redo button, 29:object detection button, 30:label volume and annotation time display panel, 31:user comment delivery hyperlink)

Figure 3 illustrates the step-by-step process and the dialog box user interfaces for loading the images and their corresponding masks. The real time image brightness, contrast, and zoom can be adjusted by using the brightness, contrast, and zoom scroll buttons shown in numbers 23, 24, and 25 in Figure 2. Additionally, the view axis can be changed using the radio buttons indicated by the number 8 in Figure 2.

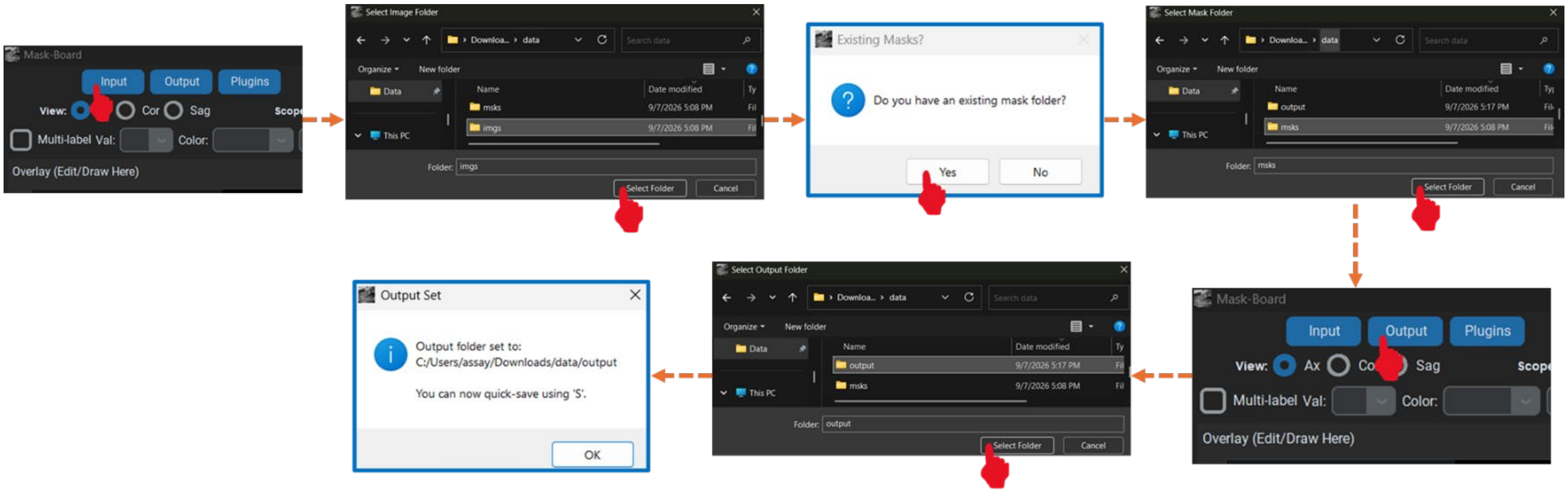

Figure 3: The workflow for selecting the directory for loading image batch, initial prediction batch, selecting output destination folder.

The manual annotation was done with the MedImgAnnotator tool using a structured, semi-automated workflow. In each case the corresponding volumetric scan was loaded and identified by file name and index in the annotation queue, and Prev/Next (showed by the numbers 5, 6, and 7 in Figure 2) controls were used to navigate between cases. Figure 2 illustrates the buttons, search bars, scroll bars, and viewing panels of the annotation tool. The choice of the viewing plane (Axial, Coronal or Sagittal) and annotation scope (2D for single slice editing or 3D for volumetric editing) was made depending on the anatomical structure to be segmented. For cases with multiple structures annotated in the same volume, the multi-label option was enabled, and each structure was given a unique label value and display color to preserve the same color-coding across slices and cases. The segmentation was then performed using one of four tools, depending on the morphology of the structure. The Brush tool allows freehand pixel-wise delineation of the structure. The Bucket tool allows flood-fill labeling of contiguous regions. The Polygon tool allows defining the boundaries of the structure by placing vertices (with an optional automatic-closure mode). The Wand tool allows intensity-based region selection with a tolerance threshold and pixel-count limit. Annotators navigated through the volume using the slice slider and corrected boundaries if necessary, using the Eraser tool. For multi-slice structures, annotation was performed on a sparse subset of representative slices to reduce manual workload, and the interpolation function was used to automatically generate intermediate-slice masks between the reference annotations. During the process the overlay opacity was modified to visually check the mask alignment with the underlying image intensity and any incorrect edits were reverted using the undo function. When a case was fully annotated and visually verified, the resulting mask was saved to disk and the annotator moved on to the next case in the queue. For further elaboration, the video guide given by the link MedImgAnnotator guide - YouTube can be used.

### 3.2. Image preparation and processing package

The integrated preprocessing toolbox incorporates the data format conversion, image resampling and reorientation, intensity normalization, image enhancement, connected component analysis and dataset curation modules. The module facilitates image preparation and preprocessing tasks prior to annotation. Figure 4 shows the overview of ImPrep package.

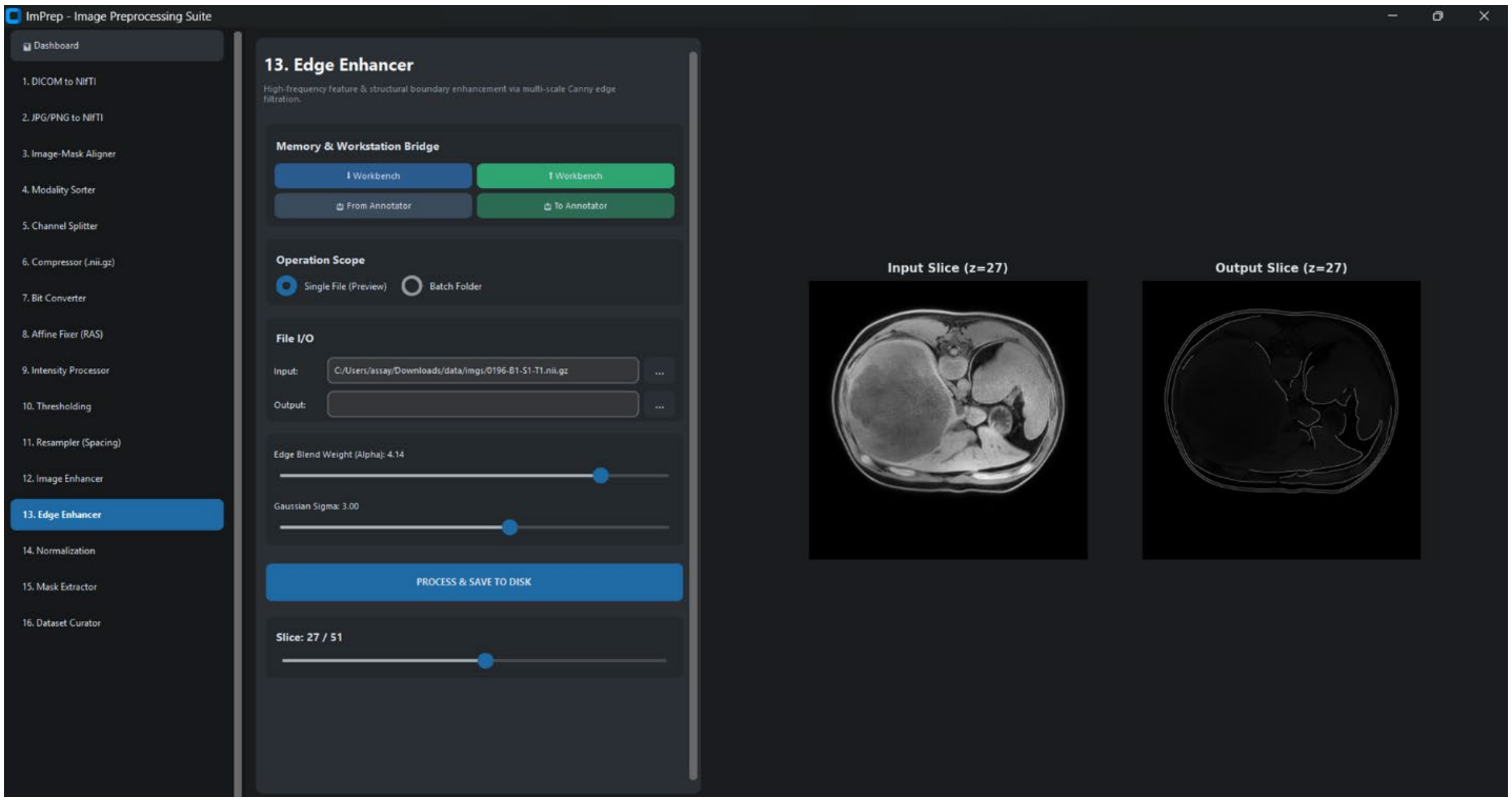

Figure 4: the overview of the image preparation and preprocessing toolbox

### 3.3. Plugins

The developed annotation and preprocessing tool incorporated a plugin manager that allows the user to install, delete, check the files in the plugin folder, and run the installed plugin application from the MedImgAnnotator user interface. Figure 5 shows the steps that the user has to follow to install and use the plugins. One automated deep learning-based liver segmentation model and one image preparation and preprocessing packages were incorporated with the developed annotation and preprocessing tool. Figure 6 shows the user interface of the two incorporated plugins.

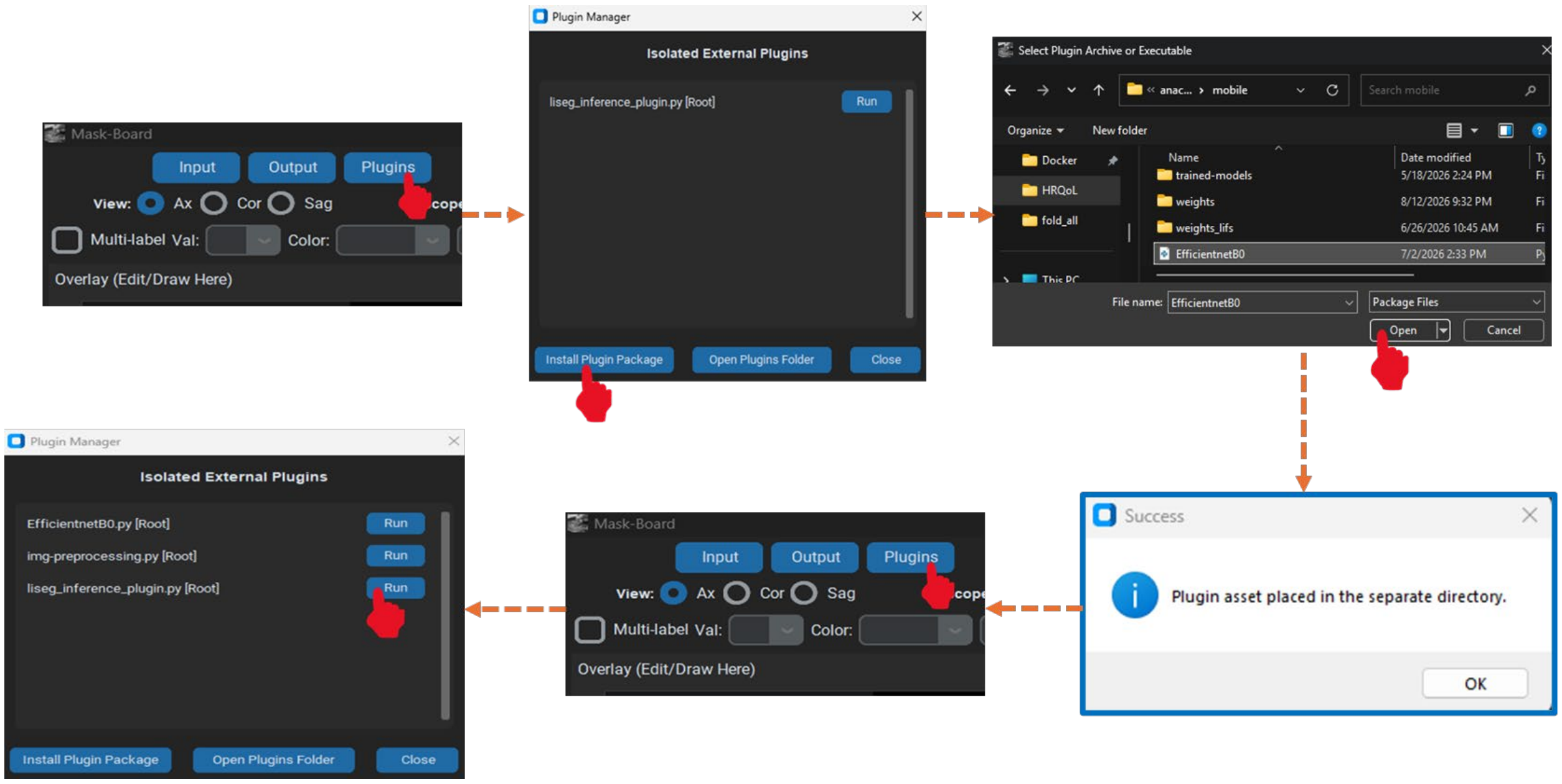

Figure 5: plugin installation and utilization procedure

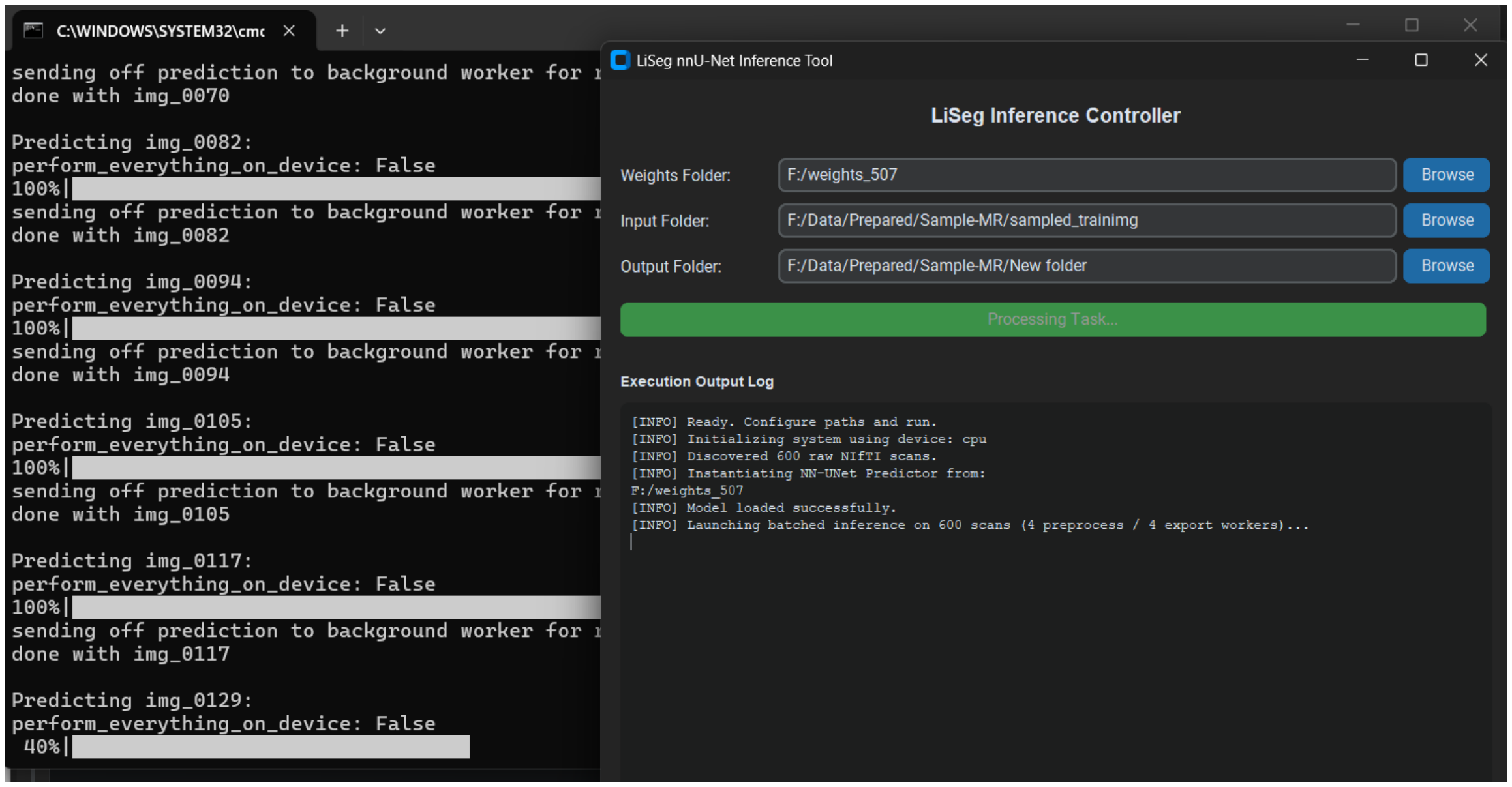

Figure 6: Graphical user interface overview of automated liver segmentation plugins.

## 3.4. Evaluating the Tool

### 3.4.1. Comparing the tool with openly available annotation tools

MedImgAnnotator was able to reduce the active annotation time required to segment liver from abdominal volume to 59 seconds, which is 2.55–3.95-fold lower than the required active annotation time by MRIcron, ITK-SNAP, 3D

Slicer, and ImageJ. Table 1 and Figure 7 show the active annotation time it took for segmenting liver using five annotation tools.

Table 1: Comparison of active annotation time per slice across five medical image annotation tools.

| Tasks | Annotation tool (required time in seconds) | | | | |
|---|---|---|---|---|---|
| | MRIcron (seconds) | ITK-snap (seconds) | 3D-Slicer (seconds) | ImageJ (seconds) | MedImgAnnotator (seconds) |
| Image opening | 11 | 22 | 14 | 22 | 7 |
| Creating mask | 60 | 80 | 120 | 60 | 30 |
| Erasing mask | 60 | 60 | 60 | 80 | 10 |
| Undo | 5 | - | - | 7 | - |
| Saving mask | 9 | 28 | 20 | 40 | 5 |
| Opening and overlaying mask | 13 | 15 | 17 | 35 | 7 |
| Overall single image process | 153 | 205 | 231 | 237 | 59 |
| Opening image batch | NA | NA | 90 | NA | 3 |
| Image-mask batch opening and matching | NA | NA | NA | NA | 3 |

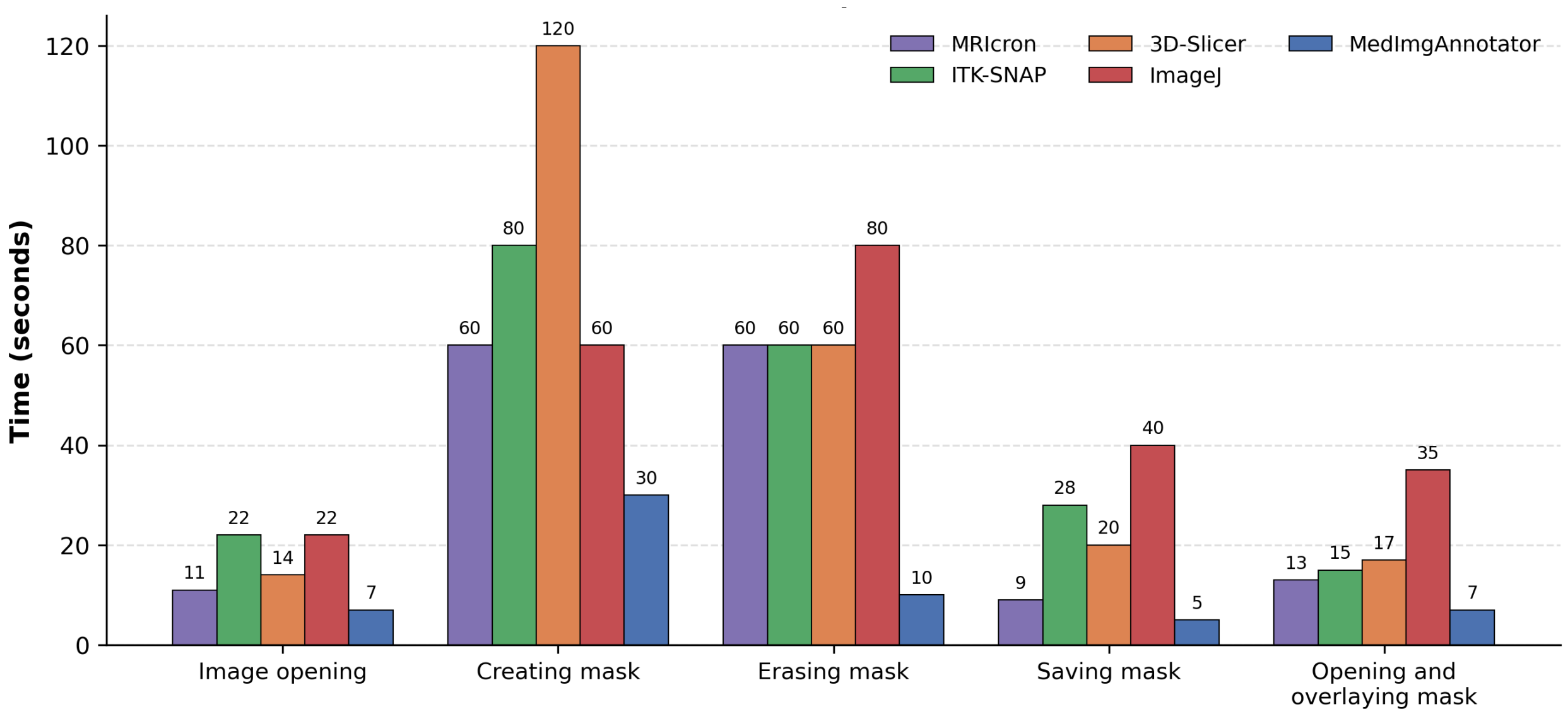

Figure 7: comparison of active annotation time across five annotation tools

The performed Friedman Test resulted in a chi-square statistic (Q) of 15.53 and a p-value of 0.00372, showing that MedImgAnnotator significantly reduces the required annotation time compared to the commonly used openly available annotation tools.

### 3.4.2. User testing and comparison results with their favorite annotation tools

Usability testing with 12 independent medical imaging researchers confirmed that MedImgAnnotator significantly reduced curation time and interaction overhead compared to preferred baseline tools as shown in Table 2, and Figure 8. MedImgAnnotator shortened active annotation time by 32.0%. User-reported simplicity on a 5-point Likert scale was also significantly higher for MedImgAnnotator with 11 of 12 researchers rating it equal to or better than their baseline software. The detailed user evaluation results are presented in Table 2 and Figure 8.

Table 2: The comparative evaluation results of users with respect to their routinely used annotation tool

| User-ID | Modality and ROI | Routinely used annotation tool | | | MedImgAnnotator | |
|---|---|---|---|---|---|---|
| | | Name | Time (Sec) | Rating | Time (Sec) | Rating |
| 1 | MRI, Liver | ImageJ | 480 | 3 | 360 | 4 |
| 2 | MRI, Liver | 3D-Slicer | 770 | 4 | 796 | 4 |
| 3 | CT, Lung | 3D-Slicer | 496 | 3 | 367 | 5 |
| 4 | CT, Lungs | 3D-Slicer | 350 | 5 | 120 | 5 |
| 5 | MRI, Liver | MRIcron | 600 | 4 | 125 | 5 |
| 6 | Lung Nodules | MRIcron | - | 4 | - | 4 |
| 7 | MRI, TUMOR | ITK-SNAP | 120 | 3 | 65 | 3 |
| 8 | CT, Coronary artery | 3D-Slicer | 65 | 4 | 60 | 5 |
| 9 | MRI, Brain Tumor | 3D-Slicer | 290 | 4 | 72 | 5 |
| 10 | CT, multi-organ | 3D-Slicer | 900 | 2 | 300 | 4 |
| 11 | MRI, Brain tumor | ImageJ | 359 | 4 | 318 | 5 |
| 12 | MRI, Brain tumor | 3D-Slicer | 190 | 4 | 350 | 5 |
| | | Mean | 420.00 | 3.67 | 266.64 | 4.50 |
| | | Standard deviation | 262.51 | 0.78 | 217.37 | 0.67 |
| | | Batch labeling support | Not available | | Available | |

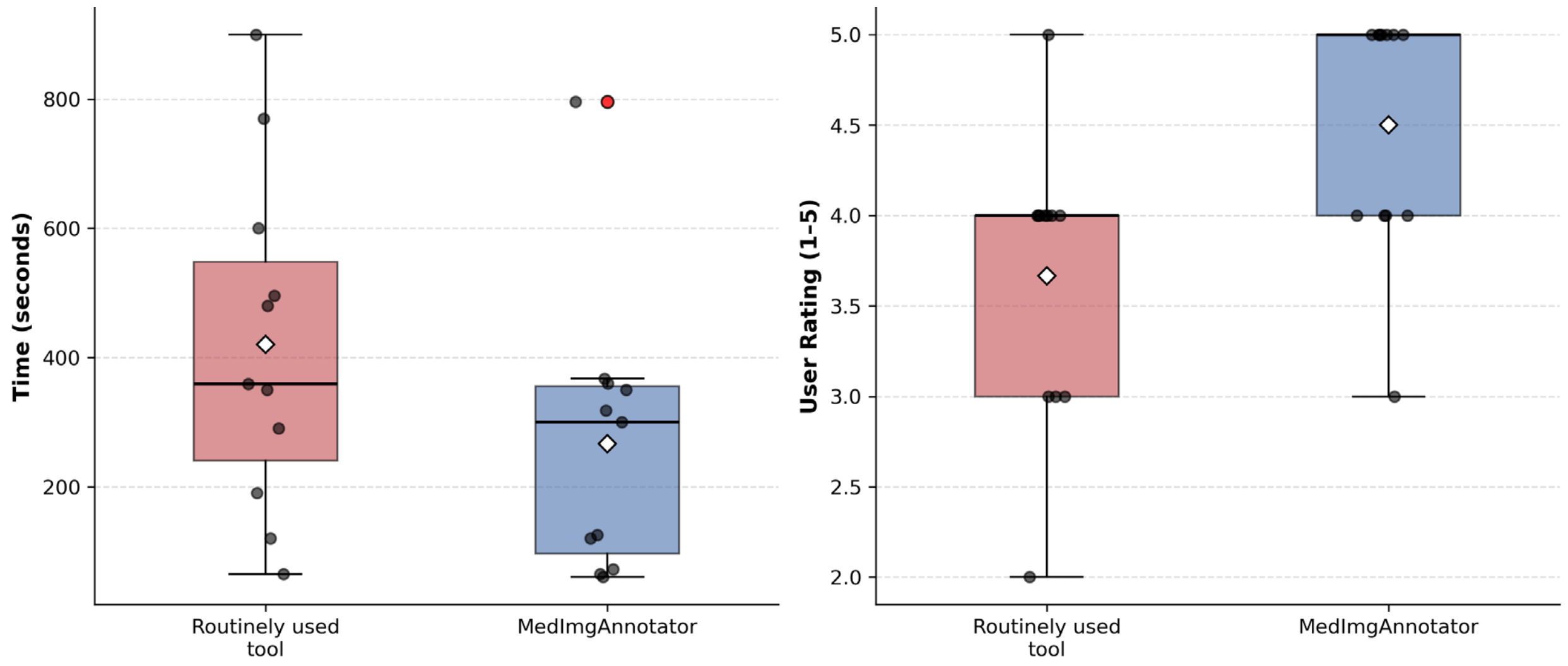

Figure 8: The comparison of ImageJ, 3D-slicer, MRIcron, and ITK-snap against MedImgAnnotator in terms of annotation time and user rating.

MedImgAnnotator reduced annotation time by 36.5% compared with users' routinely used tools. Across the 11 users paired data, the mean annotation time was 266.64 ± 217.37 seconds (median 300 s; IQR 96–355 s) for MedImgAnnotator, versus 420.00 ± 262.51 seconds (median 359 s; IQR 240–548 s) for the routinely used tool. A two-sided Wilcoxon signed-rank test showed that this reduction was statistically significant ($W = 9$, $p = 0.032$), with nine of the 11 users faster with MedImgAnnotator. User-rated ease of use was also higher for MedImgAnnotator (4.50 ± 0.67 vs 3.67 ± 0.78; median 5 [IQR 4–5] vs 4 [IQR 3–4]; $n = 12$; Wilcoxon $p = 0.008$).

### 3.4.3. Annotating different regions of interests

MedImgAnnotator maintained rapid curation and minimal user input across diverse anatomical structures and modalities as shown in the Tables 3–5 and Figure 9. Median liver segmentation times were 95.4 s on abdominal CT and 22.4 s on abdominal MR (n = 20 each). Two CT outliers (229.9–339.9 s) were driven by severe cirrhosis, ascites, and multifocal metastases requiring manual slice-by-slice editing. Across other organs, mean active annotation time ranged from 25.1 s for the pancreas and 55.3 s for lungs to 315.2 s for complex brain tumors, confirming high efficiency across varying anatomical complexities.

Table 3: Active annotation time for liver parenchyma segmentation across 20 CT and 20 MR volumetric examinations

| **CT scans** | | **MR Scans** | |
|---|---|---|---|
| Image_Name | Time_Elapsed_Sec | Image_Name | Time_Elapsed_Sec |
| patient1 | 215 | img_0001 | 43.1 |
| patient10 | 101.1 | img_0002 | 47.8 |
| patient11 | 30.3 | img_0003 | 147.8 |
| patient12 | 78.2 | img_0004 | 16.3 |
| patient13 | 185.6 | img_0005 | 4.4 |
| patient14 | 89.7 | img_0006 | 5.8 |
| patient15 | 124.9 | img_0007 | 28.5 |
| patient16 | 13.5 | img_0008 | 113.7 |
| patient17 | 79.7 | img_0009 | 62.9 |
| patient18 | 40.8 | img_0010 | 7.7 |
| patient19 | 69.4 | img_0011 | 54.1 |
| patient20 | 161.4 | img_0012 | 62.8 |
| patient2 | 21.3 | img_0013 | 6 |
| patient3 | 207.2 | img_0014 | 8.8 |
| patient4 | 54 | img_0015 | 15.5 |
| patient5 | 101.8 | img_0016 | 29.9 |
| patient6 | 146.4 | img_0017 | 57.1 |
| patient7 | 339.9 | img_0018 | 7.4 |
| patient8 | 13.6 | img_0019 | 102.3 |
| patient9 | 229.9 | img_0020 | 11.6 |
| Mean | 115.19 | Mean | 41.68 |
| STDV | 86.44 | STDV | 40.63 |

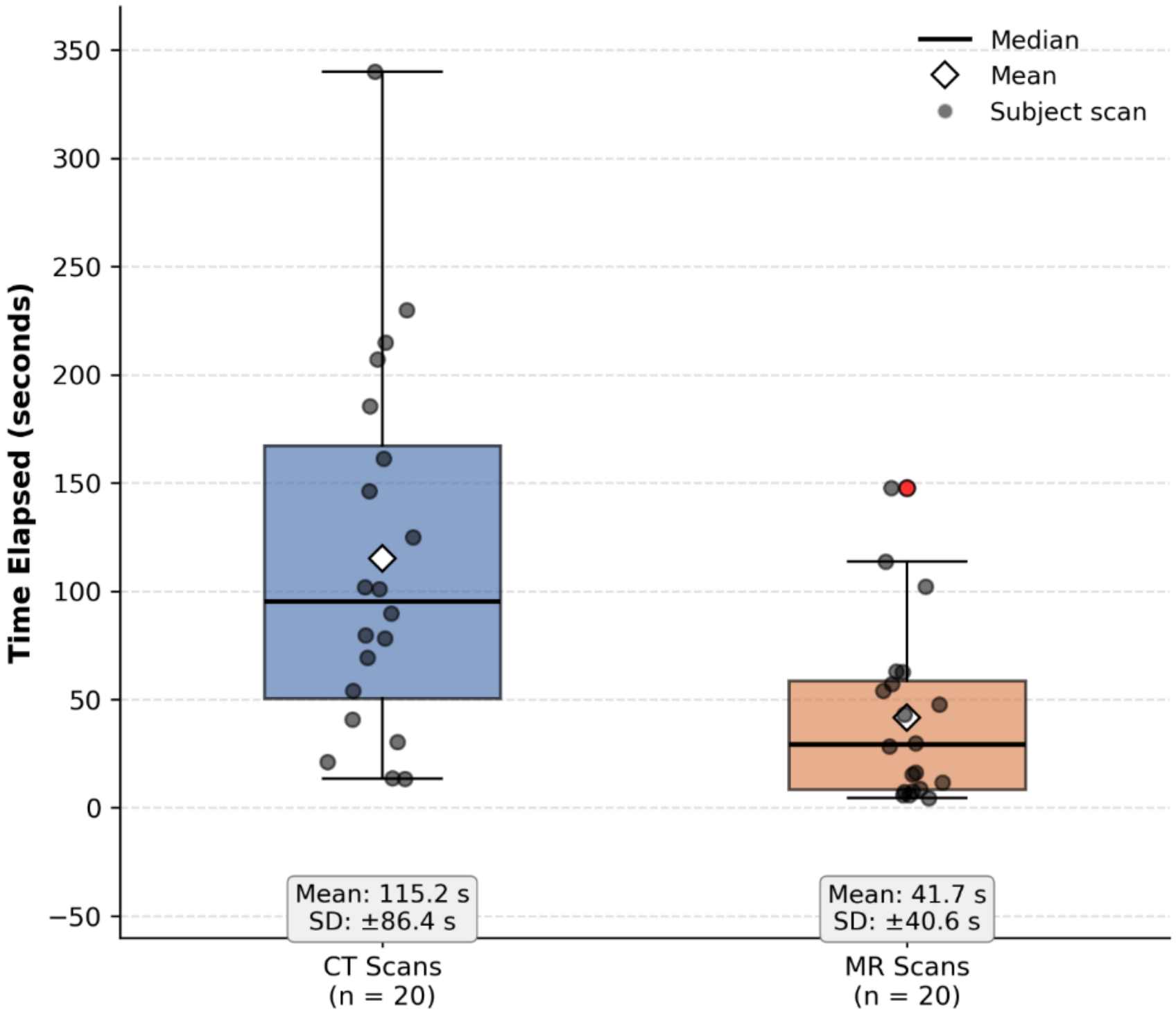

Figure 9: Distribution of active annotation time for liver parenchyma segmentation across 20 CT and 20 MR volumetric examinations

The average active annotation time of MedImgAnnotator for liver parenchyma segmentation on 20 abdominal CT volumes was 115.19±86.44 s (median: 95.40 s; IQR: 47.40–173.50 s). In abdominal MR volumes (n=20), the active annotation time was significantly shorter (41.68 ± 40.63 s, median: 22.40 s, IQR: 7.55–60.00 s). The Mann-Whitney U test result shows that the annotation time required for MR scans is significantly lower compared to CT scans with a p-value of 0.00148.

Table 4: Active annotation time for brain tumor, lung parenchyma, pancreas, spleen, and kidney segmentation

| Brain MR scans | Chest MR scans | Abdominal MR scans | | | |
|---|---|---|---|---|---|
| Brain tumor segmentation | Lung segmentation | Organ of interest | Pancreas | Spleen | Kidney |

| Image_Name | TES | Image_Name | TES | Image_Name | TES | TES | TES |
|---|---|---|---|---|---|---|---|
| P1w1CT1 | 474.7 | img_0016 | 33.3 | AHN_0001_0000 | 51.7 | 45.6 | 77.2 |
| P2wOCT1 | 128.2 | img_0017 | 35 | AHN_0002_0000 | 21.3 | 77.7 | 66.2 |
| P3w0CT1 | 344.7 | img_0018 | 39.9 | AHN_0003_0000 | 12.3 | 46.6 | 21.6 |
| P4wOCT1 | 213.6 | img_0019 | 43.8 | AHN_0004_0000 | 21.6 | 44.9 | 37.3 |
| P5wOCT1 | 414.9 | img_0020 | 124.7 | AHN_0005_0000 | 18.7 | 36.5 | 25.3 |
| Mean | 315.22 | Mean | 55.34 | Mean | 25.12 | 50.26 | 45.52 |
| STDV | 142.78 | STDV | 38.99 | STDV | 15.32 | 15.86 | 24.9 |

The Kruskal-Walli's test result shows that there is a significant difference in the annotation time between brain tumor and lung parenchyma annotation from five brain MR scans and 5 chest MR scans, with a Kruskal–Wallis $H = 6.82$, and $p = 0.009$. Among the abdominal organs, the mean annotation time was $25.12 \pm 15.32$ s for the pancreas, $50.26 \pm 15.86$ s for the spleen, and $45.52 \pm 24.90$ s for the kidney. The Friedman test showed no statistically significant difference among the three organs (chi square (Q) = 5.20, $p = 0.074$), although the pancreas was the fastest organ to annotate in four of the five volumes.

## 4. Discussion

Preparing good quality ground-truth labels for thousands of clinical volumetric scans is widely acknowledged as the primary bottleneck in medical imaging artificial intelligence [2, 8]. Conventional open-access biomedical imaging software packages such as such as 3D Slicer [9], ITK-SNAP [10, 11], ImageJ/Fiji [12], MITK [13], and MRIcron [14] were designed comprehensive image analysis, simulation, and prototyping. However, prolonged annotation times, repetitive manual navigation, and the lack of automated image–label matching substantially limit the utility of these tools. MedImgAnnotator was specifically designed to overcome these bottlenecks, maximizing operator throughput by integrating automated directory traversal, name similarity-based string matching, front-panel editing controls, slice interpolation, and one-click saving into a lightweight, standalone desktop application. The results of this study substantiate the marked temporal advantages of MedImgAnnotator. A single-case baseline evaluation showed that MedImgAnnotator reduced the active annotation time from the case baseline evaluation. MedImgAnnotator reduced active annotation time from 206.5±32.87 seconds to 59 seconds.

Moreover, Wilcoxon signed-rank test on the comparative annotation time evaluation of MedImgAnnotator with four commonly used open-source annotation tools shows a significant ($p = 0.032$) reduction in active annotation time. Additionally, a significantly higher (0.0078) ease-of-use rating was obtained from MedImgAnnotator, as rated by 12 users, compared to that of four commonly used open-source annotation tools. The annotation test performed on CT and MR scans shows that the annotation time required for CT is higher compared to MR scans, as CT has a commonly larger number of slices. Annotation of brain tumors, lungs, pancreas, spleen, and kidneys was performed using the current tool, and the annotation time varies depending on the complexity and size of the region of interest.

According to the literature, commonly used annotation tools such as 3D Slicer[9], ITK-SNAP[10, 11], ImageJ/Fiji[12], MITK[13], and MRIcron[14] offer powerful algorithmic backends but suffer from procedural complexity, prolonged time consumption, and an inability to natively support automated image-mask batch processing[47, 48]. 3D Slicer, ITK-SNAP, ImageJ/Fiji, MITK, and MRIcron require repetitive and tedious data loading, image mask pairing and overlaying, editing, and saving procedures. MedImgAnnotator addresses the time-consuming, repetitive, and tedious gaps of the openly available annotation tools by significantly reducing the required active annotation time with its minimal semiautomated interaction functions and batch matching algorithms for image masks. Additionally, it provides a streamlined end-to-end workflow by integrating image cleaning, preparation, preprocessing, dataset curation, and image annotation (segmentation and detection) functions into a single system.

Despite its simplicity and efficiency, the current tool has limitations: The executable file is restricted to the Windows operating system; it cannot show multiparameter and multiphasic scans side by side, and it lacks 3D visualization of the volumes. Furthermore, the tool was tested by only 12 users, which needs to be tested by a larger cohort of participants from different centers and professions for better generalizability.

## 5. Conclusion

The proposed MedImgAnnotator is a simple and efficient semi-automated batch image annotation and preprocessing tool aiming to reduce time-consuming and tedious annotation procedures. The tool utilizes automated batch queuing, name similarity-based image-mask pair matching, front-panel editing tools, and slice interpolation techniques that significantly reduced the active annotation time by 32.0% compared to the openly available annotation tools. Thus, this standalone executable and comprehensive tool provide a better option for researchers to perform annotation on larger-sized data.

## 6. Statements & Declarations

### 6.1. Funding

The authors did not receive any funds for conducting and publishing this study

### 6.2. Competing Interests

An informed written consent form was obtained from all study participants

### 6.3. Author Contributions

Ewunate Assaye Kassaw conceptualized, designed, and implemented this study in collaboration with the co-authors professor Anup Singh and professor Amit Mehndiratta. Doctor Peush Sahni reviewed the study from a clinical research usability perspective. All authors contributed to the design, prototyping, and testing of the study. The article was drafted by Ewunate Assaye Kassaw and reviewed by Professor Anup Singh, Professor Amit Mehndiratta, and Doctor Peush Sahni. All co-authors had the opportunity to comment on the manuscript and approved the final version for publication.

### 6.4. Ethics approval

This study did not require ethical approval, as it did not involve the direct recruitment of human participants or the collection of new patient data. All imaging data used for the development and evaluation of MedImgAnnotator were obtained from publicly available, de-identified datasets, including the AMOS (Abdominal Multi-Organ Segmentation) dataset, the Pancreas MRI dataset, and the 3D Slicer sample template data, all of which are freely accessible for research purposes under their respective open-access licenses. As these datasets are fully anonymized and were not collected specifically for this study, no additional ethical clearance or institutional review board (IRB) approval was necessary. Furthermore, the individuals who utilized the tool to evaluate its performance were medical image processing researchers assessing software usability and annotation efficiency, not patients or human research subjects; therefore, informed consent and ethics committee review were not applicable to this component of the study.

### 6.5. Consent for publication

This research did not involve humans, animals, or other subjects

### 6.6. Acknowledgments

Resources required to conduct the study were provided by the MedImg lab at the Indian Institute of Technology Delhi. The members of the MedImg lab were helpful in testing the developed tool as users.

## Reference

1. Litjens G, Kooi T, Bejnordi BE, et al (2017) A survey on deep learning in medical image analysis. Med Image Anal 42:60–88. https://doi.org/10.1016/j.media.2017.07.005

2. Gao Y, Jiang Y, Peng Y, et al (2025) Medical Image Segmentation: A Comprehensive Review of Deep Learning-Based Methods. Tomography 11:. https://doi.org/10.3390/tomography11050052

3. Ma J, He Y, Li F, et al (2024) Segment anything in medical images. Nat Commun 15:654. https://doi.org/10.1038/s41467-024-44824-z

4. Zou X, Yang J, Zhang H, et al (2023) Segment Everything Everywhere All at Once

5. Ma J, He Y, Li F, et al (2024) Segment anything in medical images. Nat Commun 15:654. https://doi.org/10.1038/s41467-024-44824-z

6. Zhu J, Hamdi A, Qi Y, et al (2024) Medical SAM 2: Segment medical images as video via Segment Anything Model 2

7. Ravi N, Gabeur V, Hu Y-T, et al (2024) SAM 2: Segment Anything in Images and Videos

8. Tajbakhsh N, Jeyaseelan L, Li Q, et al (2020) Embracing imperfect datasets: A review of deep learning solutions for medical image segmentation. Med Image Anal 63:101693. https://doi.org/10.1016/j.media.2020.101693

9. Fedorov A, Beichel R, Kalpathy-Cramer J, et al (2012) 3D Slicer as an image computing platform for the Quantitative Imaging Network. Magn Reson Imaging 30:1323–1341. https://doi.org/10.1016/j.mri.2012.05.001

10. Yushkevich PA, Pashchinskiy A, Oguz I, et al (2019) User-Guided Segmentation of Multi-modality Medical Imaging Datasets with ITK-SNAP. Neuroinformatics 17:83–102. https://doi.org/10.1007/s12021-018-9385-x

11. Yushkevich PA, Piven J, Hazlett HC, et al (2006) User-guided 3D active contour segmentation of anatomical structures: Significantly improved efficiency and reliability. NeuroImage 31:1116–1128. https://doi.org/10.1016/j.neuroimage.2006.01.015

12. Schindelin J, Arganda-Carreras I, Frise E, et al (2012) Fiji: an open-source platform for biological-image analysis. Nat Methods 9:676–682. https://doi.org/10.1038/nmeth.2019

13. Wolf I, Vetter M, Wegner I, et al (2005) The Medical Imaging Interaction Toolkit. Med Image Anal 9:594–604. https://doi.org/10.1016/j.media.2005.04.005

14. Rorden C (2025) From MRIcro to MRIcron: The evolution of neuroimaging visualization tools. Neuropsychologia 207:109067. https://doi.org/10.1016/j.neuropsychologia.2025.109067

15. Yushkevich PA, Gao Y, Gerig G (2016) ITK-SNAP: An interactive tool for semi-automatic segmentation of multi-modality biomedical images. In: 2016 38th Annual International Conference of the IEEE Engineering in Medicine and Biology Society (EMBC). pp 3342–3345

16. Xu X, Wang T, Shi Y, et al (2019) Whole Heart and Great Vessel Segmentation in Congenital Heart Disease Using Deep Neural Networks and Graph Matching. In: Medical Image Computing and Computer Assisted Intervention – MICCAI 2019: 22nd International Conference, Shenzhen, China, October 13–17, 2019, Proceedings, Part II. Springer-Verlag, Berlin, Heidelberg, pp 477–485

17. Xu X (2025) XiaoweiXu/Whole-heart-and-great-vessel-segmentation-of-chd_segmentation

18. Ji Y, Bai H, Ge C, et al AMOS: A Large-Scale Abdominal Multi-Organ Benchmark for Versatile Medical Image Segmentation

19. Zhang Z, Keles E, Durak G, et al (2024) Large-Scale Multi-Center CT and MRI Segmentation of Pancreas with Deep Learning. In: arXiv.org. https://arxiv.org/abs/2405.12367v3. Accessed 21 Feb 2026

20. Kassaw EA, Mehndiratta A, Sahni P, Singh A (2026) Iterative Human-in-the-Loop Learning for Accurate Segmentation of Liver

21. Ratcliff JW, Metzener, DE (1988) Pattern Matching: The Gestalt Approach. https://jacobfilipp.com/DrDobbs/articles/DDJ/1988/8807/8807c/8807c.htm. Accessed 30 July 2026

22. Schwarzhans F, George G, Escudero Sanchez L, et al (2025) Image normalization techniques and their effect on the robustness and predictive power of breast MRI radiomics. Eur J Radiol 187:112086. https://doi.org/10.1016/j.ejrad.2025.112086

23. Wang CJ, Rost NS, Golland P (2023) Spatial-Intensity Transforms for Medical Image-to-Image Translation. IEEE Trans Med Imaging 42:3362–3373. https://doi.org/10.1109/TMI.2023.3283948

24. Gonzalez RC, Woods RE (2018) Digital image processing. Pearson, New York, NY

25. Yalcin H, Collins R, Hebert M (2007) Background estimation under rapid gain change in thermal imagery. Comput Vis Image Underst 106:148–161. https://doi.org/10.1016/j.cviu.2006.10.007

26. Porter T, Duff T (1984) Compositing digital images. In: Proceedings of the 11th annual conference on Computer graphics and interactive techniques. Association for Computing Machinery, New York, NY, USA, pp 253–259

27. Wolberg G (1998) Image morphing: a survey. Vis Comput 14:360–372. https://doi.org/10.1007/s003710050148

28. Lehmann TM, Gonner C, Spitzer K (1999) Survey: interpolation methods in medical image processing. IEEE Trans Med Imaging 18:1049–1075. https://doi.org/10.1109/42.816070

29. Williams L (1983) Pyramidal parametrics. ACM SIGGRAPH Comput Graph 17:1–11. https://doi.org/10.1145/964967.801126

30. Keys R (1981) Cubic convolution interpolation for digital image processing. IEEE Trans Acoust Speech Signal Process 29:1153–1160. https://doi.org/10.1109/TASSP.1981.1163711

31. Oliphant TE (2007) Python for Scientific Computing. Comput Sci Eng 9:10–20. https://doi.org/10.1109/MCSE.2007.58

32. Sung-Hyuk C Comprehensive Survey on Distance/Similarity Measures between Probability Density Functions. Int J Math Models Methods Appl Sci. https://doi.org/https://www.naun.org/main/NAUN/ijmmas/mmmas-49.pdf

33. Hughes JF Computer Graphics: Principles and Practice

34. Chaikin GM (1974) An algorithm for high-speed curve generation. Comput Graph Image Process 3:346–349. https://doi.org/10.1016/0146-664X(74)90028-8

35. De Berg M, Cheong O, Van Kreveld M, Overmars M (2008) Computational Geometry: Algorithms and Applications. Springer Berlin Heidelberg, Berlin, Heidelberg

36. He L, Chao Y, Suzuki K, Wu K (2009) Fast connected-component labeling. Pattern Recognit 42:1977–1987. https://doi.org/10.1016/j.patcog.2008.10.013

37. Falcao AX, Stolfi J, De Alencar Lotufo R (2004) The image foresting transform: theory, algorithms, and applications. IEEE Trans Pattern Anal Mach Intell 26:19–29. https://doi.org/10.1109/TPAMI.2004.1261076

38. Lotufo RA, Falcao AX, Zampirolli FA (2002) IFT-Watershed from gray-scale marker. In: Proceedings. XV Brazilian Symposium on Computer Graphics and Image Processing. pp 146–152

39. Raya SP, Udupa JK (1990) Shape-based interpolation of multidimensional objects. IEEE Trans Med Imaging 9:32–42. https://doi.org/10.1109/42.52980

40. Ye Q-Z (1988) The signed Euclidean distance transform and its applications. In: [1988 Proceedings] 9th International Conference on Pattern Recognition. pp 495–499 vol.1

41. Strutz T (2023) The Distance Transform and its Computation

42. Cuisenaire O, Macq B (1999) Fast and exact signed Euclidean distance transformation with linear complexity. In: 1999 IEEE International Conference on Acoustics, Speech, and Signal Processing. Proceedings. ICASSP99 (Cat. No.99CH36258). pp 3293–3296 vol.6

43. Wang H, Dong L, O'Daniel J, et al (2005) Validation of an accelerated 'demons' algorithm for deformable image registration in radiation therapy. Phys Med Biol 50:2887. https://doi.org/10.1088/0031-9155/50/12/011

44. Thirion J-P (1998) Image matching as a diffusion process: an analogy with Maxwell's demons. Med Image Anal 2:243–260. https://doi.org/10.1016/S1361-8415(98)80022-4

45. Marr D, Hildreth E (1980) Theory of edge detection. Proc R Soc Lond B Biol Sci 207:187–217. https://doi.org/10.1098/rspb.1980.0020

46. Isensee F, Jaeger PF, Kohl SAA, et al (2021) nnU-Net: a self-configuring method for deep learning-based biomedical image segmentation. Nat Methods 18:203–211. https://doi.org/10.1038/s41592-020-01008-z

47. Maier-Hein L, Eisenmann M, Reinke A, et al (2018) Why rankings of biomedical image analysis competitions should be interpreted with care. Nat Commun 9:5217. https://doi.org/10.1038/s41467-018-07619-7

48. Jha S, Topol EJ (2016) Adapting to Artificial Intelligence: Radiologists and Pathologists as Information Specialists. JAMA 316:2353. https://doi.org/10.1001/jama.2016.17438